\documentclass[english]{scrartcl}
\usepackage[T1]{fontenc}
\usepackage[utf8]{inputenc}
\usepackage{babel}
\usepackage{mathptmx}
\usepackage{subcaption}
\usepackage{graphicx}
\usepackage{gensymb}
\usepackage{tabu}
\usepackage{multirow}
\usepackage{siunitx}
\usepackage[
    backend=biber,
    style=ieee,
    sorting=none
]{biblatex}
\usepackage{url}
\usepackage{hyperref}

\usepackage{tikz}
\usepackage{pgfplots}
\pgfplotsset{compat=newest}
\usepgfplotslibrary{fillbetween, groupplots}
\tikzset{font={\fontsize{14pt}{12}\selectfont}}

\newcommand{\todo}[1]{\iffalse #1 \fi}

\setkomafont{author}{\normalsize}
\setkomafont{disposition}{\normalfont\bfseries}
\setkomafont{title}{\large\bfseries}

\title{Decentralised Semi-supervised Onboard Learning for Scene Classification in Low-Earth Orbit
}

\author{%
{\bfseries\large Johan \"{O}stman$^{1}$, Pablo G\'{o}mez$^{1,3}$, Vinutha Magal Shreenath  $^1$, Gabriele Meoni$^{1,2}$}\\
\small\\
$^1$ AI Sweden, Lindholmspiren 11, 417 56 Göteborg, Sweden \\
\\
$^2$ $\Phi$-Lab, European Space Agency, ESRIN, Frascati, Italy\\
\\
$^3$ Advanced Concepts Team, European Space Agency, ESTEC, Noordwijk, The Netherlands\\
\href{mailto:{pablo.gomez}@esa.int}{pablo.gomez@esa.int}}
\date{}

\todo{make repo public}
\todo{Show the average over satellites}

\begin{document}
\maketitle

\begin{abstract}
Onboard machine learning on the latest satellite hardware offers the potential for significant savings in communication and operational costs. We showcase the training of a machine learning model on a satellite constellation for scene classification using semi-supervised learning while accounting for operational constraints such as temperature and limited power budgets based on satellite processor benchmarks of the neural network. We evaluate mission scenarios employing both decentralized and federated learning approaches. All scenarios achieve convergence to high accuracy (around 91\% on EuroSAT RGB dataset) within a one-day mission timeframe.
\end{abstract}

\section{Introduction}

A new generation of satellites is currently bringing hardware suitable for machine learning (ML) onboard spacecraft into Earth orbit. 
Recent works \cite{matthiesen2022federated} explored the possibility to train ML models in a distributed manner onboard satellite constellations. 
Distributed onboard training brings the potential to reduce communication requirements, operational cost and time, and improve autonomy by sharing ML models, trained close to the sensors, instead of the collected data.
While previous missions have demonstrated the ability to perform inference onboard spacecraft for data processing \cite{GiuffridaPhiSat}, training onboard presents additional challenges. Convincingly addressing operational constraints is crucial, as the computational cost of training is significantly higher, and the lack of labeled examples during the mission can often be prohibitive. 

In this work, we investigate the training of an ML model onboard a satellite constellation for scene classification. We employ a semi-supervised learning approach called MSMatch \cite{Gomez21}, which we successfully distribute using decentralized learning techniques. Operational constraints such as temperature, communication windows, and limited power budgets are modeled using PASEOS \cite{Gomez23}, a specialized Python module. We provide detailed results on various scenarios involving decentralized and federated learning approaches.

\section{Methods}
This work is built upon three core components: the semi-supervised learning method MSMatch, modeling constraints with PASEOS, and adapting MSMatch for distributed implementation.

\subsection{MSMatch} 
One of the primary challenges for ML applications on spacecraft is the scarcity of labeled training data, particularly before launch. Often, there is an insufficient number of labeled examples for training a model on the ground, necessitating the use of semi- and self-supervised techniques in many instances. MSMatch is a semi-supervised classification method specifically designed for such scenarios \cite{Gomez21}, and has been proven to achieve high accuracy even when trained with merely a few labels per class.

MSMatch employs consistency regularization and pseudo-labeling to train primarily on unlabeled images. It fundamentally relies on the consistency between the model's predictions on two differently augmented (one strongly, one weakly) versions of the same image. A pseudo-label is generated for the weakly augmented version. Additionally, a supervised loss is applied to the limited available labeled examples. With as few as five labeled samples per class, MSMatch can achieve accuracies above 90\% on established benchmarks such as EuroSAT \cite{Helber}. The method has also demonstrated effectiveness with multispectral data.

For our implementation, we built upon the existing open-source codebase available online\footnote{\url{https://github.com/gomezzz/MSMatch} Accessed: 2023-02-27}. We utilized the EfficientNet-lite models (efficientnet-lite0), derived from the original EfficientNets \cite{tan2019efficientnet}, as the backend. Due to their small memory footprint and efficiency, these models are well-suited for embedded systems and, therefore, onboard processors.

\subsection{PASEOS}
Training machine learning models in space necessitates accounting for factors such as power budgets, thermal management, and communication windows, as these directly impact the viability of training \cite{Gomez23}. Communication windows, in particular, are a critical factor in distributed computing scenarios \cite{matthiesen2022federated,razmi2022board}.

To model these constraints, we employ the open-source Python module PASEOS \cite{Gomez23} (Version~0.1.3). PASEOS simulates spacecraft orbital dynamics and power budgets, taking into account power consumption, available solar panels, and eclipses. Thermal management is modeled using a single-node ordinary differential equation \cite{Gomez23}. Packet communication is calculated based on the assumed available bandwidth and the presence of a line of sight between communication partners. PASEOS operates asynchronously to the training pipeline, thereby limiting the ability to train and exchange models. A comprehensive description of the models can be found in the article dedicated to PASEOS~\cite{Gomez23}.

\subsection{Decentralized MSMatch}

\begin{figure}[h]
\centering
      \includegraphics[width=\textwidth]{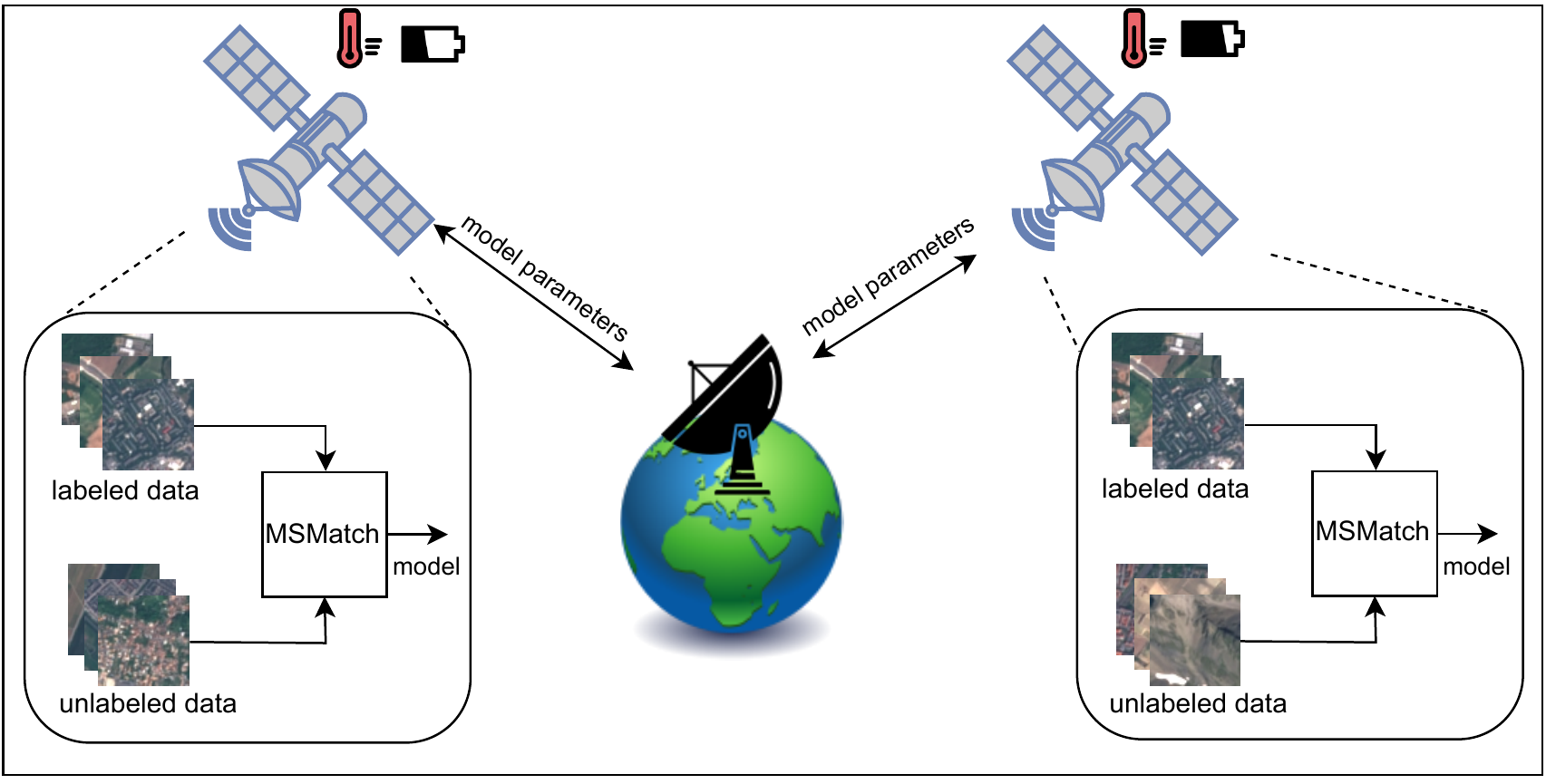}
      \caption{Decentralized MSMatch for two satellites and a ground station.}
      \label{fig:dec_msmatch}
\end{figure}

In this study, we demonstrate the capability of training MSMatch in a distributed environment by leveraging well-established techniques for merging local models, such as federated averaging~\cite{matthiesen2022federated}. The Message Passing Interface (MPI) is employed to enable asynchronous training of multiple models while concurrently running PASEOS simulations for each satellite.
It is assumed that labeled training data are available prior to launch and preloaded onto each satellite. Unlabeled training data, on the other hand, are randomly distributed among the satellites, with each satellite receiving a fixed number of distinct, unlabeled samples. The hyperparameters of the decentralized MSMatch are comparable to those in~\cite{Gomez21}, with a few exceptions: the batch size is reduced to 32 for labeled data and 96 for unlabeled data, and the learning rate is increased to 0.03. MPI facilitates communication if PASEOS simulations indicate an available window for data exchange.
A decentralized MSMatch scenario involving two satellites and a ground station is depicted in Fig.~\ref{fig:dec_msmatch}. It is important to note that while the satellites share the same labeled examples, the unlabeled data differ. The code for our work is openly accessible online\footnote{\url{https://github.com/gomezzz/DistMSMatch} Accessed: 2023-04-05}.

\section{Results}
\subsection{Setup and Scenarios}
To test the proposed method we rely on the EuroSAT dataset~ \cite{Helber} used in the original MSMatch paper~\cite{Gomez21} to enable a direct comparison. 
The dataset is comprised of 27000 $64 \times 64$ pixel, 13-channel images from Sentinel-2A data classified into ten classes, such as forest or river.
In our experimental results, we utilize only the RGB channels.
The choice to employ solely the RGB channels stems from two factors: firstly, the RGB channels have already demonstrated satisfactory performance; and secondly, the inclusion of all channels would substantially prolong the training time. However, it should be noted that relying on only the RGB channels places us in a less favorable situation, and leveraging more available data would likely enhance performance further.

To demonstrate the ability to learn in a semi-supervised way, five labeled images from each class, i.e., 50 labeled images in total, are extracted and loaded onto each satellite. 
That is, the satellites are loaded with the same 50 images.
The test set consists of 2700 images and the remaining 24250 images are treated as unlabeled data and are randomly split into eight partitions, i.e., 3031.25 images per partition on average.

The satellites utilize a radio frequency (RF) link of 1 Mbps when communicating with a ground station, and optical inter-satellite links (ISL)  in orbit at 100 Mbps. 
To account for tracking and alignment, we assume the ISL between two satellites to exhibit a setup-time of 30 seconds before every communication attempt~\cite{Bhattacharjee2023}.
The satellites are equipped with a 0.2772 MJ battery and solar panels that charge at 20 W.
The parameters of the thermal model assume a mass of six kilograms, an initial temperature of 26.85 degree Celsius, sun absorptance of 0.9, and an infrared absorptance of 0.5. 
The sun-facing and Earth-facing area of each satellite are 0.015~m$^2$ and 0.01~m$^2$, respectively. The emissive area is 0.1~m$^2$ and the thermal capacity is 5000 J/(kg * K).
The utilized EfficientNet-lite0 network occupies 12.7 MB of storage in a compressed state and the model exchange with the ground takes 201.78 seconds whereas the model exchange via ISL takes 32.03 seconds.
Training a batch required 15.98 seconds on a Unibap iX10 satellite processor CPU.
Further, communications (to ground), communications (ISL), training, and standby are assumed to consume 10 W, 13.5 W, 30 W, and 5 W, respectively.
The PASEOS simulation is run with the default configuration of v0.1.3. The initial epoch of the simulation is 2022-Dec-17 14:42:42.

\begin{figure}[h]
    \begin{subfigure}[b]{0.32\textwidth}
      \includegraphics[width=\textwidth]{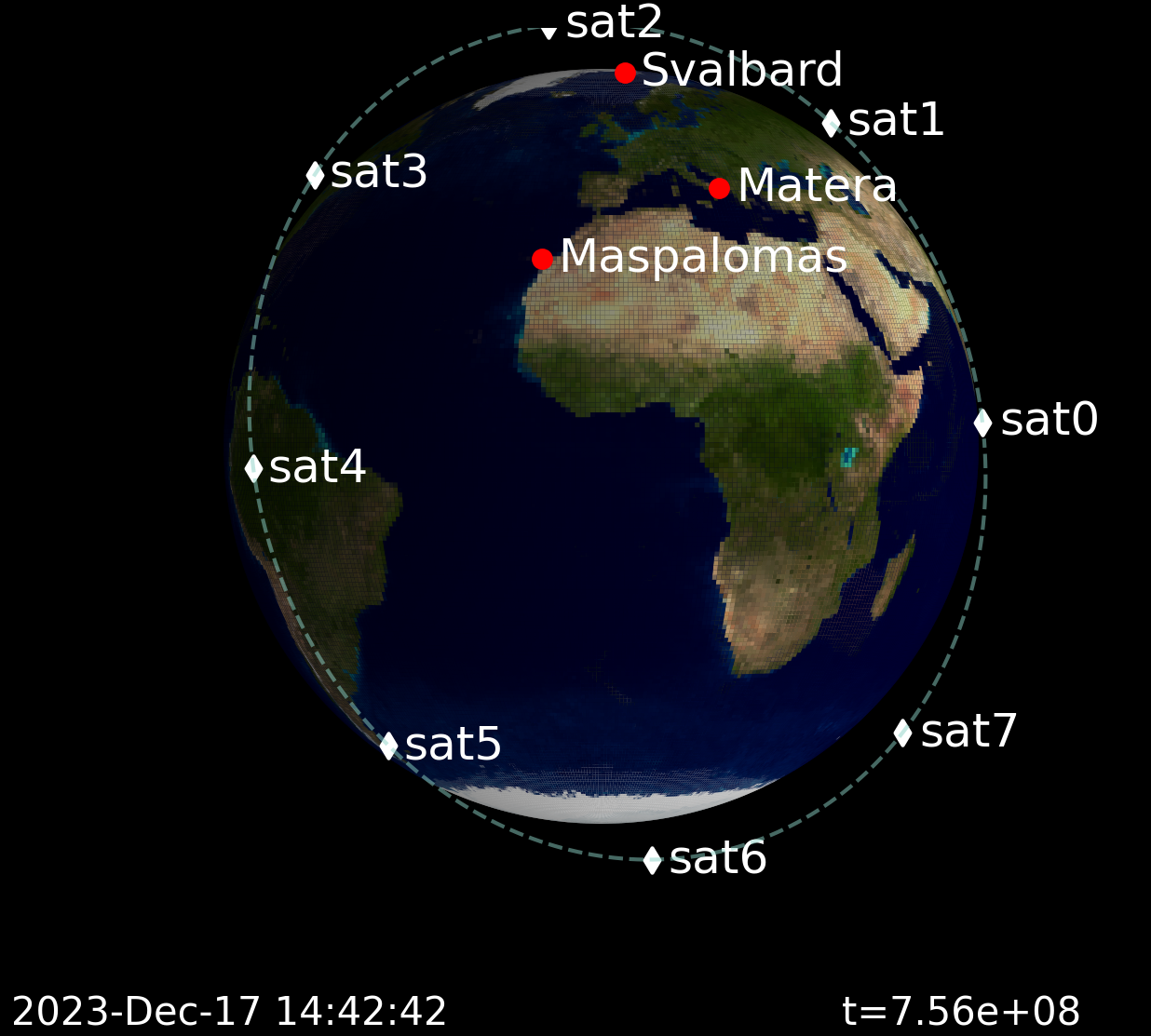}
      \caption{\textsc{Ground Station}.}
      \end{subfigure}
    \begin{subfigure}[b]{0.32\textwidth}
      \includegraphics[width=\textwidth]{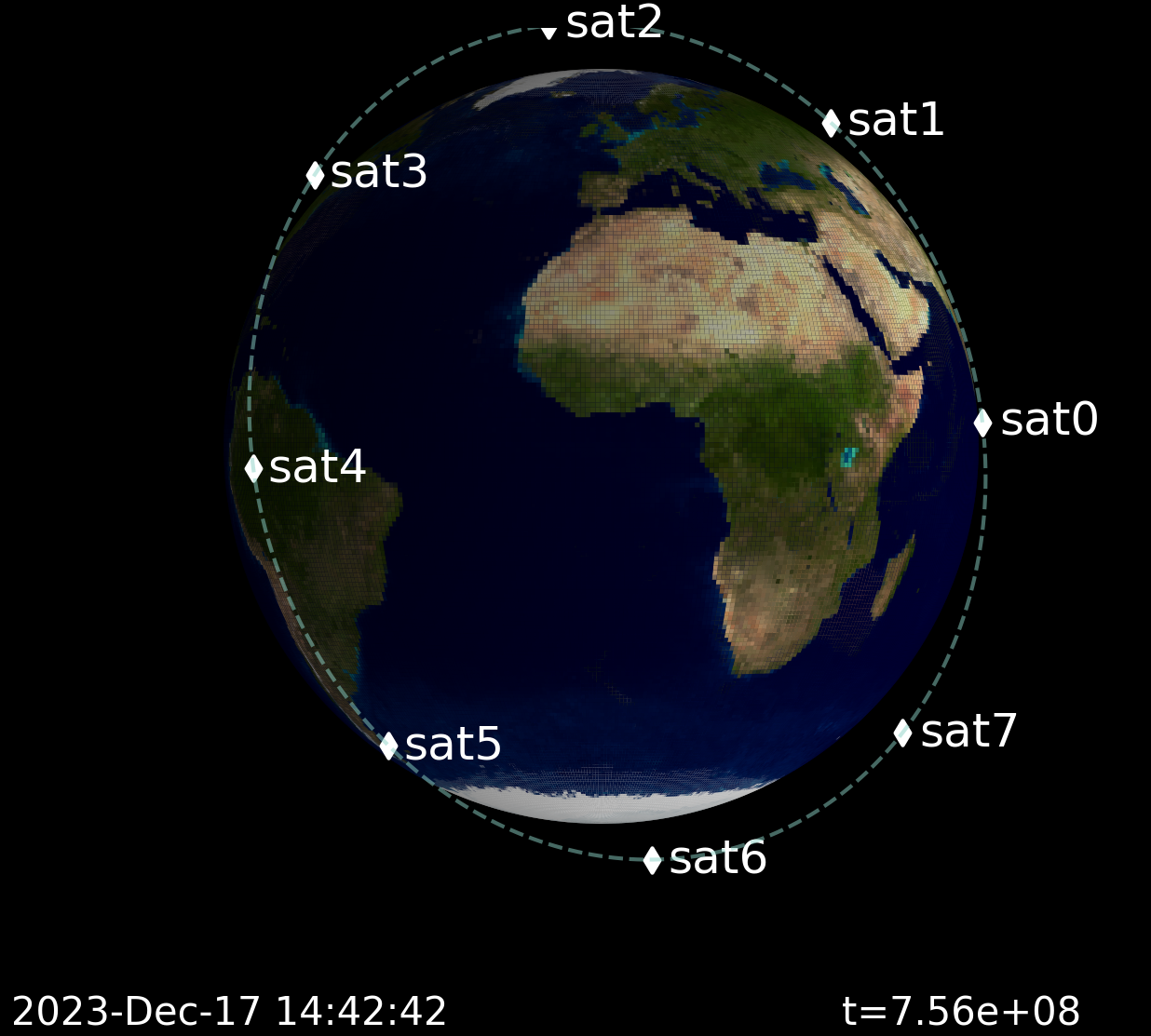}
      \caption{\textsc{Swarm}.}
    \end{subfigure}
    \begin{subfigure}[b]{0.32\textwidth}
      \includegraphics[width=\textwidth]{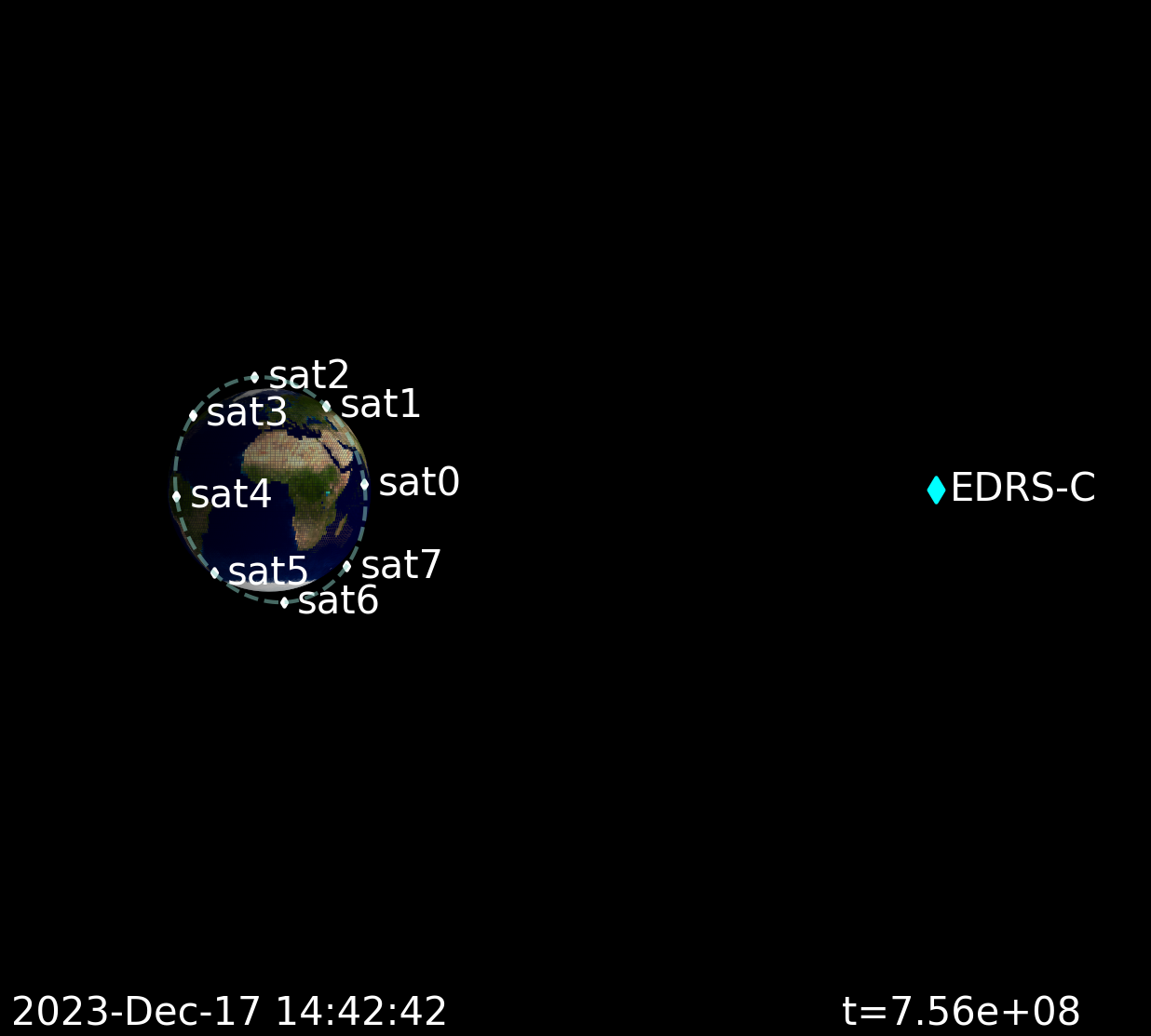}
      \caption{\textsc{Relay}.}
    \end{subfigure}
    \caption{Visualization of the different constellations and communication setups.}
    \label{fig:scenarios}
\end{figure}

The investigated, distributed scenarios, as displayed in Fig.~\ref{fig:scenarios}, involve a constellation in a Sentinel-like orbit (sun-synchronous at 786 km altitude with 98.62\degree inclination) featuring eight satellites. 
The first scenario (\textsc{Ground Station}) assumes model exchanges via three (linked) ground stations on Gran Canaria, Svalbard and in Matera, Italy.
From a satellite perspective, the ground stations are viewed as a single unit.
In the second (\textsc{Swarm}), eight satellites are assumed to communicate via ISL. 
Finally, in the third (\textsc{Relay}), one of the European Data Relay Satellite System relays (EDRS-C) is assumed to act as a central server for federated learning. Communication delays due to the relay potentially being busy are neglected.
In the federated settings (\textsc{Ground Station} and \textsc{Relay}), the global model is updated asynchronously whenever a local model becomes available by a convex combination, with weight 0.4, of the global model and the newly received model similarly to~\cite{xie2019asynchronous} (with constant weighting function).
Furthermore, each of the satellites will attempt to share their local models every 1000 seconds.

\subsection{Training Results}

\begin{table*}[b]\scriptsize
\caption{Results averaged over the eight satellites over three different runs. \label{table:main_results}}
\begin{center}
 {\tabulinesep=1.2mm
 \setlength\tabcolsep{2pt}
\begin{tabu} { | X[0.5cm] | X[0.60cm] | X[0.5cm] | X[0.5cm] | X[0.55cm] | X[0.7cm] | X[0.45cm] |}
\hline
Setup & Accuracy [\%] &  Transmitted Data [MB] & Power Consumption [Wh] & Time Training [\% of total] & Time Communicating [\% of total] & time between communications [s] \\\hline
\textsc{Ground Station}   & 91.51 $\pm 0.95$   & 185.74 & 447.58  & 54.25 & 1.71 & 5913.6  \\ \hline
\textsc{Swarm}  & 90.96$\pm 1.34$  & 1168.40 & 449.36 & 53.76 & 3.34  & 1878.4\\ \hline
\textsc{Relay}  & 91.19$\pm 0.76$ & 455.61 & 449.37 & 54.45 & 1.32 & 2349.3\\ \hline
\end{tabu}}
\end{center}
\end{table*}

The results obtained from 24 hours of simulation time (equivalent to 14.34 orbital revolutions) have been averaged over three independent runs per scenario and are presented in Table~\ref{table:main_results}. All three scenarios attain an average accuracy exceeding 91\%, with the \textsc{Ground Station} scenario achieving the highest accuracy at 91.51\%.
It can also be seen that the standard deviation is the lowest for \textsc{Relay} and largest for \textsc{Swarm}. 
This is expected as the \text{Relay} and \textsc{Ground Station} scenarios involves sharing a global model in contrast to the \textsc{Swarm} scenario. 

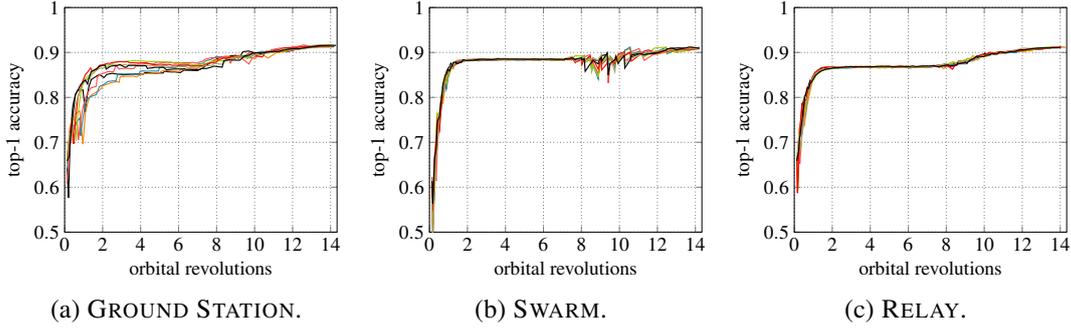
\begin{figure}[t]
    \begin{subfigure}[b]{0.32\textwidth}
    \resizebox{!}{0.8\linewidth}{
      \begin{tikzpicture}
\begin{axis}
[
    grid = both, 
    grid style={dotted,draw=black!90},
    tick label style={/pgf/number format/fixed},
    xmode = linear, 
    ymode = linear, 
    ymax = 1, 
    ymin = 0.5, 
    xmax = 14.34, 
    xmin = 0, 
    xlabel =  orbital revolutions, 
    ylabel =  top-1 accuracy, 
    cycle list name=exotic,
]

    \addplot[thick, solid, mark=none] table [x index = {0}, y index={2}, col sep=comma]
    {./data/FL_groundnode0_accuracy.csv}; 
    \addplot+[thick, solid,mark=none] table [x index = {0}, y index={2}, col sep=comma]
    {./data/FL_groundnode1_accuracy.csv}; 
    \addplot+[thick, solid, mark=none] table [x index = {0}, y index={2}, col sep=comma]
    {./data/FL_groundnode2_accuracy.csv}; 
    \addplot+[thick, solid, mark=none] table [x index = {0}, y index={2}, col sep=comma]
    {./data/FL_groundnode3_accuracy.csv}; 
    \addplot+[thick, solid, mark=none] table [x index = {0}, y index={2}, col sep=comma]
    {./data/FL_groundnode4_accuracy.csv}; 
    \addplot+[thick, solid, mark=none] table [x index = {0}, y index={2}, col sep=comma]
    {./data/FL_groundnode5_accuracy.csv}; 
    \addplot+[thick, solid, mark=none] table [x index = {0}, y index={2}, col sep=comma]
    {./data/FL_groundnode6_accuracy.csv}; 
    \addplot+[thick, solid, mark=none] table [x index = {0}, y index={2}, col sep=comma]
    {./data/FL_groundnode7_accuracy.csv}; 
\end{axis}
\end{tikzpicture}
      }
      \caption{\textsc{Ground Station}.}
      \end{subfigure}
    \begin{subfigure}[b]{0.32\textwidth}
    \resizebox{!}{0.8\linewidth}{
      \begin{tikzpicture}
\begin{axis}
    [
    grid = both, 
    grid style={dotted,draw=black!90},
    tick label style={/pgf/number format/fixed},
    xmode = linear, 
    ymode = linear, 
    ymax = 1, 
    ymin = 0.5, 
    xmax = 14.34, 
    xmin = 0, 
    xlabel =  orbital revolutions, 
    ylabel =  top-1 accuracy, 
    cycle list name=exotic,
    ]

    \addplot[thick, solid, mark=none] table [x index = {0}, y index={2}, col sep=comma]
    {./data/Swarmnode0_accuracy.csv}; 
    \addplot+[thick, solid, mark=none] table [x index = {0}, y index={2}, col sep=comma]
    {./data/Swarmnode1_accuracy.csv}; 
    \addplot+[thick, solid, mark=none] table [x index = {0}, y index={2}, col sep=comma]
    {./data/Swarmnode2_accuracy.csv}; 
    \addplot+[thick, solid, mark=none] table [x index = {0}, y index={2}, col sep=comma]
    {./data/Swarmnode3_accuracy.csv}; 
    \addplot+[thick, solid, mark=none] table [x index = {0}, y index={2}, col sep=comma]
    {./data/Swarmnode4_accuracy.csv}; 
    \addplot+[thick, solid, mark=none] table [x index = {0}, y index={2}, col sep=comma]
    {./data/Swarmnode5_accuracy.csv}; 
    \addplot+[thick, solid, mark=none] table [x index = {0}, y index={2}, col sep=comma]
    {./data/Swarmnode6_accuracy.csv}; 
    \addplot+[thick, solid, mark=none] table [x index = {0}, y index={2}, col sep=comma]
    {./data/Swarmnode7_accuracy.csv}; 
\end{axis}

\end{tikzpicture}}
      \caption{\textsc{Swarm}.}
    \end{subfigure}
    \begin{subfigure}[b]{0.32\textwidth}
    \resizebox{!}{0.8\linewidth}{
      \begin{tikzpicture}
\begin{axis}
    [
    grid = both, 
    grid style={dotted,draw=black!90},
    tick label style={/pgf/number format/fixed},
    xmode = linear, 
    ymode = linear, 
    ymax = 1, 
    ymin = 0.5, 
    xmax = 14.34, 
    xmin = 0, 
    xlabel =  orbital revolutions, 
    ylabel =  top-1 accuracy, 
    cycle list name=exotic,
    ]

    \addplot[thick, solid, mark=none] table [x index = {0}, y index={2}, col sep=comma]
    {./data/FL_geostatnode0_accuracy.csv}; 
    \addplot+[thick, solid, mark=none] table [x index = {0}, y index={2}, col sep=comma]
    {./data/FL_geostatnode1_accuracy.csv}; 
    \addplot+[thick, solid, mark=none] table [x index = {0}, y index={2}, col sep=comma]
    {./data/FL_geostatnode2_accuracy.csv}; 
    \addplot+[thick, solid, mark=none] table [x index = {0}, y index={2}, col sep=comma]
    {./data/FL_geostatnode3_accuracy.csv}; 
    \addplot+[thick, solid, mark=none] table [x index = {0}, y index={2}, col sep=comma]
    {./data/FL_geostatnode4_accuracy.csv}; 
    \addplot+[thick, solid, mark=none] table [x index = {0}, y index={2}, col sep=comma]
    {./data/FL_geostatnode5_accuracy.csv}; 
    \addplot+[thick, solid, mark=none] table [x index = {0}, y index={2}, col sep=comma]
    {./data/FL_geostatnode6_accuracy.csv}; 
    \addplot+[thick, solid, mark=none] table [x index = {0}, y index={2}, col sep=comma]
    {./data/FL_geostatnode7_accuracy.csv}; 
\end{axis}
\end{tikzpicture}
    }
      \caption{\textsc{Relay}.}
    \end{subfigure}
    \caption{Top-1 accuracy on the test set for the satellites averaged over three runs. Different colors indicate different satellites.}
    \label{fig:results}
\end{figure}

The \textsc{Swarm} scenario is the most communication-intensive, with satellites sharing an average of 1168.40 MB of data. This is because there are always neighboring satellites available to receive local models. In contrast, satellites in the \textsc{Relay} scenario transmit an average of 455.61 MB of data, which is considerably less than in the \textsc{Swarm} scenario. This difference is due to the relay satellite occasionally being obscured by Earth. The \textsc{Ground Station} scenario has the least data transmission, with satellites transmitting only 185.74 MB of data, as the link to the ground stations is infrequently available.

Over the simulated 24-hour period, satellites in the \textsc{Ground Station}, \textsc{Swarm}, and \textsc{Relay} scenarios communicate their local models an average of 14.625, 46, and 35.875 times, respectively. It is important to note, however, that this is not reflected in the relative time spent communicating, as the \textsc{Relay} scenario spends the least time communicating due to the \textsc{Ground Station} relying on RF communications. The power consumption and total training time are similar across all three scenarios.
Note, however, that the cost of operating the ground station or the relay satellite is not accounted for.

The convergence of the top-1 accuracy (averaged over three independent runs) for each satellite is depicted in Fig.~\ref{fig:results}. The performance among different satellites in the \textsc{Swarm} and \textsc{Relay} scenarios is more consistent due to frequent communication, which prevents satellites from deviating significantly. The \textsc{Swarm} scenario reaches a local optimum after approximately two orbital revolutions, resulting in an 88.5\% top-1 accuracy, while the \textsc{Relay} scenario attains a less favorable local optimum in the same time frame, with an 86.7\% top-1 accuracy. Interestingly, satellites in both scenarios escape their local optima after eight orbital revolutions and find a more favorable optimum after 12 revolutions. Although the \textsc{Ground Station} scenario is not as consistent as the other two scenarios, it exhibits similar behavior.

\begin{figure}[tb]
\centering
    \begin{subfigure}[b]{0.3\textwidth}
    \resizebox{!}{0.8\linewidth}{
      \begin{tikzpicture}

\begin{axis}[
  ylabel style={blue},
  yticklabel style={blue},
  axis y line*=left,
  ymin=295, ymax=325,
  xmin=0, xmax=14.34,
  restrict y to domain=295:325, 
  xlabel=orbital revolutions,
  ylabel={temperature [K]}
]
    \addplot[blue, thick, solid, mark=none] table [x index = {0}, y index={1}, col sep=comma]
    {./data/FL_ground_paseos_data.csv}; 
    \addplot[blue, dashed, mark=none] table [x index = {0}, y index={7}, col sep=comma]
    {./data/FL_ground_paseos_data.csv}; 

\end{axis}

\begin{axis}[
  ylabel ={},
  yticklabels =none,
  axis y line*=right,
  axis x line=none,
  ymin=0.15, ymax=0.55,
    xmin=0, xmax=14.34,
  restrict y to domain=0.15:0.55, 
]
    \addplot[red, thick, solid, mark=none] table [x index = {0}, y index={4}, col sep=comma]
    {./data/FL_ground_paseos_data.csv}; 
    \addplot[red, dashed, mark=none] table [x index = {0}, y index={8}, col sep=comma]
    {./data/FL_ground_paseos_data.csv}; 

\end{axis}

\end{tikzpicture}
      }
      \caption{\textsc{Ground Station}.}
      \end{subfigure}
      \hspace{0.32cm}
    \begin{subfigure}[b]{0.3\textwidth}
    \resizebox{!}{0.8\linewidth}{
      \begin{tikzpicture}

\begin{axis}[
 ylabel ={},
  yticklabels = none,
  axis y line*=none,
  ymin=295, ymax=325,
  xmin=0, xmax=14.34,
  restrict y to domain=295:325, 
  xlabel=orbital revolutions,
]
    \addplot[blue, thick, solid, mark=none] table [x index = {0}, y index={1}, col sep=comma]
    {./data/Swarm_paseos_data.csv}; 
    \addplot[blue, dashed, mark=none] table [x index = {0}, y index={7}, col sep=comma]
    {./data/Swarm_paseos_data.csv}; 

\end{axis}

\begin{axis}[
 ylabel ={},
   yticklabels = none,
  axis y line*=right,
  axis x line=none,
  ymin=0.15, ymax=0.55,
    xmin=0, xmax=14.34,
  restrict y to domain=0.15:0.55, 
]
    \addplot[red, thick, solid, mark=none] table [x index = {0}, y index={4}, col sep=comma]
    {./data/Swarm_paseos_data.csv}; 
    \addplot[red, dashed, mark=none] table [x index = {0}, y index={8}, col sep=comma]
    {./data/Swarm_paseos_data.csv}; 

\end{axis}

\end{tikzpicture}}
      \caption{\textsc{Swarm}.}
    \end{subfigure}
    \begin{subfigure}[b]{0.3\textwidth}
    \resizebox{!}{0.8\linewidth}{
      \begin{tikzpicture}

\begin{axis}[
 ylabel ={},
   yticklabels = none,
  axis y line*=none,
  ymin=295, ymax=325,
  xmin=0, xmax=14.34,
  restrict y to domain=295:325, 
  xlabel=orbital revolutions,
]
    \addplot[blue, thick, solid, mark=none] table [x index = {0}, y index={1}, col sep=comma]
    {./data/FL_geostat_paseos_data.csv}; 
    \addplot[blue, dashed, mark=none] table [x index = {0}, y index={7}, col sep=comma]
    {./data/FL_geostat_paseos_data.csv}; 

\end{axis}

\begin{axis}[
  ylabel style={red},
  yticklabel style={red},
  axis y line*=right,
  axis x line=none,
  ymin=0.15, ymax=0.55,
    xmin=0, xmax=14.34,
  restrict y to domain=0.15:0.55, 
  ylabel=state of charge
]
    \addplot[red, thick, solid, mark=none] table [x index = {0}, y index={4}, col sep=comma]
    {./data/FL_geostat_paseos_data.csv}; 
    \addplot[red, dashed, mark=none] table [x index = {0}, y index={8}, col sep=comma]
    {./data/FL_geostat_paseos_data.csv}; 

\end{axis}

\end{tikzpicture}}
      \caption{\textsc{Relay}.}
    \end{subfigure}
    \caption{Average temperature and state of charge of the constellation. Temperatures in blue and state of charge in red.}
    \label{fig:constraints}
\end{figure}
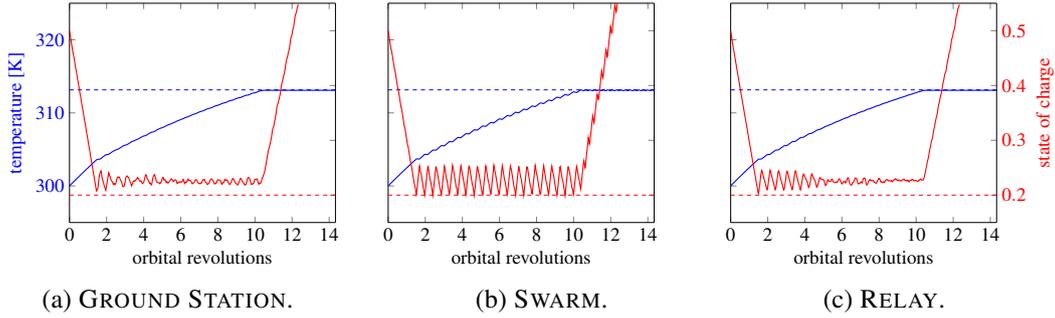

As previously mentioned, PASEOS enables accounting for constraints such as power and temperature. Figure \ref{fig:constraints} illustrates the average temperature and power consumption. In our numerical experiments, the spacecraft enters standby mode to recharge or cool down when the state of charge drops below 0.2 or the temperature exceeds 313.15 K (indicated by dashed lines in Figure \ref{fig:constraints}). However, communication is prioritized and will always be performed.

The temperature behaves similarly across all settings, as the primary influencing factor is the training process. The temperature (represented by blue curves) can be observed to trigger standby mode after approximately 10 orbital revolutions. Subsequently, the satellites enter standby mode to cool down and initiate training as soon as the temperature no longer violates the constraint, causing the temperature to oscillate around the constraint temperature.

In contrast, the state of charge (depicted by red curves) exhibits different behavior across the three scenarios. In the \textsc{Swarm} scenario, spacecraft consistently communicate their models. However, in the \textsc{Ground Station} and \textsc{Relay} scenarios, satellites do not always have a communication link available to share local models, resulting in less predictable battery consumption among the satellites.
Since training consumes the most power and the orbit allows the spacecraft to charge for most of the time, the more frequent standby mode triggered by temperature enables the spacecraft to recharge the battery.

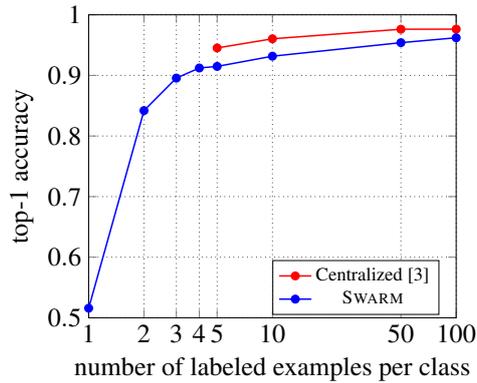
\begin{figure*}[tb]
\centering
\resizebox{!}{0.35\linewidth}{
    \begin{tikzpicture}

\begin{axis}
[
    grid = both, 
    grid style={dotted,draw=black!90},
    tick label style={/pgf/number format/fixed},
    xmode = log, 
    ymode = linear, 
    ymax = 1, 
    ymin = 0.5, 
    xmax = 100, 
    xmin = 1, 
    xlabel =  number of labeled examples per class, 
    ylabel =  top-1 accuracy, 
    cycle list name=exotic,
    xtick = {1,2,3,4,5,10,50,100},
    xticklabels={1,2,3,4,5,10,50,100},
    legend style={at={(0.5,0.1)},anchor=west}
]

    \addplot[color=red, thick, solid, mark=*] table [ x=l, y=acc, col sep=comma] {msmatch.csv};
        \addplot[color=blue, thick, solid, mark=*] table [x index = {0}, y index={1}, col sep=comma]
    {./data/Swarm_labels_vs_acc.csv}; 

    \addlegendentry{\small Centralized \cite{Gomez21}}
        \addlegendentry{\small \textsc{Swarm}}
\end{axis}

\end{tikzpicture}}
    \caption{\textsc{Swarm} top-1 test accuracy vs number of labeled examples per class. The points are averaged over satellites and three independent runs.}
    \label{fig:acc-vs-labels}
\end{figure*}

The results depicted in Figure~\ref{fig:results} are derived from 5 labeled examples per class, with the remaining data samples left unlabeled. To evaluate the influence of the labeled dataset size, we concentrate on the \textsc{Swarm} scenario and conduct the experiment with varying labeled dataset sizes. Figure~\ref{fig:acc-vs-labels} illustrates the average top-1 accuracy (across satellites and 3 independent runs).

As anticipated, the top-1 accuracy increases as the size of the labeled dataset expands. Notably, performance approaches 90\% with just 3 labeled examples per class, and with 100 labels per class, the top-1 accuracy attains 96.2\%. For comparison purposes, we include the performance of the centralized implementation of MSMatch~\cite{Gomez21}. It is important to recognize that this comparison may not be entirely fair, as the centralized version undergoes training for a substantially longer duration and employs a larger EfficientNet model. Nonetheless, the distributed version of MSMatch presented herein proves to be competitive and achieves comparable performance.

\section{Conclusion}
In this study, we illustrate the feasibility of training a state-of-the-art neural network in a semi-supervised manner, distributed across a satellite constellation using current satellite processors. Depending on the orbital configuration and assets, the constellation learns to classify the EuroSAT dataset with up to 91.51\% accuracy after a simulated training duration of 24 hours. Moving forward, the incorporation of more intricate scenarios, communication schemes, and a refined satellite architecture will enable further optimizations and increased fidelity.

\section{Acknowledgement}
The authors would like to thank Unibap AB for providing the iX-10 100 device that was used for our experiments. The work of Johan \"Ostman was funded by Vinnova under grant 2020-04825 and the work of Vinutha Magal Shreenath under Vinnova grant 2021-03643 and under Swedish National Space Agency grant 2022-00013.

\printbibliography
\end{document}